\documentclass{llncs}
\usepackage{algorithm}
\usepackage{algorithmic}
\usepackage{graphicx}
\begin{document}
\title{A Comprehensive Comparison of Pre-training Language Models}
%
%
\author{
Tong Guo
}
%

%
\institute{779222056@qq.com}
\maketitle              
\begin{abstract}

Recently, the development of pre-trained language models has brought natural language processing (NLP) tasks to the new state-of-the-art. In this paper we explore the efficiency of various pre-trained language models. We pre-train a list of transformer-based models with the same amount of text and the same training steps. The experimental results shows that the most improvement upon the origin BERT is adding the RNN-layer to capture more contextual information for short text understanding. But the conclusion is: There are no remarkable improvement for short text understanding for similar BERT structures. Data-centric method\cite{ref15} can achieve better performance.

\keywords{Deep Learning \and Pre-trained Language Model}
\end{abstract}
\section{Introduction}

In recent years, deep learning \cite{ref_proc2} and BERT \cite{ref_proc1} have shown significant improvement on almost all the NLP tasks. However, it lacks of a fair comparison on the transformer-based models, due to the pre-training datasets are different and the pre-training computing resources are different. In industry NLP applications, we need to find the most efficient BERT-like model, as the computing resources is limited.

In the paper, we pre-train a list of transformer-based models on the same datasets and pre-training steps. Then we evaluate the pre-trained performance by fine-tuning on our large text classification downstream task.

\section{Relate work}

BERT \cite{ref_proc1}, or Bidirectional Encoder Representations from Transformers, is a multi-layer transformer-encoder based \cite{ref_proc10} deep model, which produces contextual token representations that have been pre-trained from unlabeled text and fine-tuned for the supervised downstream tasks. BERT obtains state-of-the-art results on a wide array of Natural Language Processing (NLP) tasks, which include the GLUE\cite{ref_proc3} benchmark and CLUE\cite{ref_proc9} benchmark. There are two steps in BERT's framework: pre-training and fine-tuning. During pre-training, the model is trained on unlabeled data by using masked language model task and next sentence prediction task. Apart from output layers, the same architectures are used in both pre-training and fine-tuning. The same pre-trained model parameters are used to initialize models for different down-stream tasks.

\section{The Models}

In this section we describe the BERT-like models we use. First we describe the notations we use. We use $L$ to indicate the input sequence length. We use $N$ to indicate the hidden size. 

The code is released at \url{https://github.com/guotong1988/bert\_compare}

\subsection{The Origin BERT}

We refer the origin BERT to the code released by Google\footnote[1]{\url{https://github.com/google-research/bert}}.

\subsection{TextCNN-BERT}

The intuition we try this architecture is that we think convolution layer can extract features that is different from self-attention layer. We learn the model architecture from TextCNN \cite{ref_proc13}. We use TextCNN to extract feature from sequence input. Then the output of TextCNN concat the sequence input for the self-attention layer. In detail, we use a convolution kernel $R^{1 \times N}$ which in channel is $1$ and out channel is $N$ to output a tensor $R^{L \times N}$. Then we concat it to the embedding layer to get a tensor $R^{2L \times N}$ for the next self-attention-layer.

\subsection{Ngram-BERT}

The intuition we try this architecture is that we think the N-gram info can be the supplement for the one-token-level sequence. We add the N-gram info for the Origin BERT. In detail, we concat 2-gram of token embeddings to get a tensor which shape is $R^{L \times 2N}$. Then we use a matrix $R^{2N \times N}$ to transform it to a tensor $R^{L \times N}$ . Then we concat $R^{L \times N}$ to the embedding layer for the next self-attention layer.

\subsection{Dense-BERT}

The intuition we try this architecture is that we think the residual connection of transformer layers can be improved by dense connection. We learn the model architecture from DenseNet\cite{ref_proc14}. We add the dense connections in all the transformer-encoder layers. In detail, each transformer layer's input is the output of all previous layers. Although the experiment results below shows that Dense-BERT is not better than the origin BERT under the almost same parameter size, we found Dense-BERT improve the accuracy performance more as the layer number go larger.

\subsection{ConvBERT}

ConvBERT \cite{ref_proc12} is using span-based dynamic convolution to improve BERT. The code is from here\footnote[2]{\url{https://github.com/yitu-opensource/ConvBert}}.

\subsection{BORT}

BORT \cite{ref_proc11} is an optimal subarchitecture extraction for BERT by neural architecture search. We follow the final parameter setting of BORT. We only use the final parameter setting of BORT and do not use other methods proposed by the paper\cite{ref_proc11}.

\subsection{Relative Position Embedding BERT (RTE-BERT)}

We replace the embedding layer of origin BERT by the relative position embedding. The code is from here\footnote[3]{\url{https://github.com/tensorflow/tensor2tensor}}. We extract a easy-to-use relative position embedding code from tensor2tensor and put them to here\footnote[4]{\url{https://github.com/guotong1988/transformer\_relative\_position\_embedding}}.

\subsection{RNN-BERT}

We use RNN layer to capture more position info for the transformer-encoder layer. In detail, the embedding layer is followed by the LSTM layer. Then the output of LSTM layer and the embedding layer are added for the next self-attention layer. We found that concating the output of LSTM layer and the embedding layer do not get better result. 

\subsection{RNN-IN-BERT}

We use RNN layer to capture more position info for each transformer-encoder layer. In detail, the embedding layer is followed by the LSTM layer. Then the output of LSTM layer and the embedding layer are added for the next self-attention layer. The LSTM layers are also inserted between the self-attention layers.

\begin{table}
\caption{The downstream text classification evaluation results of our own dataset, under the same computing resources.}\label{tab1}
\centering
\begin{tabular}{|l|l|l|}
\hline
Model & layer setting & Accuracy \\
\hline
Origin BERT & 3 layer, 768 hidden size &  91.41\% \\ 
\hline
TextCNN-BERT & 3 layer, 768 hidden size & 91.42\% \\ 
\hline
Ngram-BERT & 3 layer, 768 hidden size & 91.32\% \\ 
\hline
Dense-BERT & 4 layer, 1024 hidden size & 91.36\% \\ 
\hline
ConvBERT\cite{ref_proc12} & 3 layer, 768 hidden size & 90.68\% \\ 
\hline
BORT\cite{ref_proc11} & 4 layer, 1024 hidden size & 91.30\% \\
\hline
RTE-BERT & 3 layer, 768 hidden size & 91.36\% \\ 
\hline
RNN-BERT & 3 layer, 768 hidden size & \textbf{91.49\%} \\ 
\hline
RNN-IN-BERT & 3 layer, 768 hidden size & \textbf{91.51\%} \\ 
\hline
\end{tabular}
\end{table}

\begin{table}
\caption{The downstream text classification evaluation results of CLUE TNEWS dataset, under the same computing resources.}\label{tab1}
\centering
\begin{tabular}{|l|l|l|}
\hline
Model & layer setting & Accuracy \\
\hline
Origin BERT & 3 layer, 768 hidden size &  54.38\% \\ 
\hline
RNN-BERT & 3 layer, 768 hidden size & 54.41\% \\ 
\hline
RNN-IN-BERT & 3 layer, 768 hidden size & \textbf{54.81\%} \\ 
\hline
\end{tabular}
\end{table}

\section{Experiments}

In this section we describe detail of experiment parameters and show the experiment results. The pre-training dataset size is 600,000,000 Chinese sentences and the downstream fine-tuning text classification dataset size is 2,000,000 Chinese sentences.
 
In pre-training, we use 400 batch size, 64 sequence length. We pre-train each kind of BERT-like model for 1,000,000 steps in the same pre-training dataset.

In fine-tuning, we use 100 batch size, 64 sequence length. We use Adam \cite{ref_proc4} with learning rate of 1e-5 and use a dropout \cite{ref_proc5} probability of 0.1 on all layers. 

For all kinds of BERT-like models, the total parameter will be no difference of 20\%.

\section{Analysis and Conclusion}

As it is shown in Table 1, we get the conclusion that the only lack of origin BERT is that the position embedding of transformer can not capture all the position or contextual info of the input sequence.

We will do more experiments on CLUE\cite{ref_proc9} in the future.


\begin{thebibliography}{2}



\bibitem{ref_proc1}
Devlin J, Chang M W, Lee K, et al. Bert: Pre-training of deep bidirectional transformers for language understanding[J]. arXiv preprint arXiv:1810.04805, 2018.

\bibitem{ref_proc2}
Krizhevsky A, Sutskever I, Hinton G E. Imagenet classification with deep convolutional neural networks[J]. Advances in neural information processing systems, 2012, 25: 1097-1105.

\bibitem{ref_proc3}
Wang A, Singh A, Michael J, et al. GLUE: A multi-task benchmark and analysis platform for natural language understanding[J]. arXiv preprint arXiv:1804.07461, 2018.

\bibitem{ref_proc4}
Kingma D P, Ba J. Adam: A method for stochastic optimization[J]. arXiv preprint arXiv:1412.6980, 2014.

\bibitem{ref_proc5}
Srivastava N, Hinton G, Krizhevsky A, et al. Dropout: a simple way to prevent neural networks from overfitting[J]. The journal of machine learning research, 2014, 15(1): 1929-1958.


\bibitem{ref_proc9}
Xu L, Hu H, Zhang X, et al. Clue: A chinese language understanding evaluation benchmark[J]. arXiv preprint arXiv:2004.05986, 2020.

\bibitem{ref_proc10}
Vaswani A, Shazeer N, Parmar N, et al. Attention is all you need[C]//Advances in neural information processing systems. 2017: 5998-6008.

\bibitem{ref_proc11}
de Wynter, Adrian, and Daniel J. Perry. "Optimal Subarchitecture Extraction For BERT." arXiv preprint arXiv:2010.10499 (2020).

\bibitem{ref_proc12}
Jiang, Zihang, et al. "Convbert: Improving bert with span-based dynamic convolution." arXiv preprint arXiv:2008.02496 (2020).

\bibitem{ref_proc13}
Zhang, Ye, and Byron Wallace. "A sensitivity analysis of (and practitioners' guide to) convolutional neural networks for sentence classification." arXiv preprint arXiv:1510.03820 (2015).

\bibitem{ref_proc14}
Huang, Gao, et al. "Densely connected convolutional networks." Proceedings of the IEEE conference on computer vision and pattern recognition. 2017.

\bibitem{ref15}
Guo T. The Re-Label Method For Data-Centric Machine Learning[J]. arXiv preprint arXiv:2302.04391, 2023.

\end{thebibliography}
\end{document}